%% file: main.tex
\documentclass[conference]{IEEEtran}
\UseRawInputEncoding
\IEEEoverridecommandlockouts
\usepackage{cite}
\usepackage{amsmath,amssymb,amsfonts}
\usepackage{algorithmic}
\usepackage{graphicx}
\usepackage{textcomp}
\usepackage{xcolor}
\def\BibTeX{{\rm B\kern-.05em{\sc i\kern-.025em b}\kern-.08em
    T\kern-.1667em\lower.7ex\hbox{E}\kern-.125emX}}
\usepackage{todonotes}

\usepackage{tikz}
\usetikzlibrary{calc}
\usetikzlibrary{shapes.misc, positioning, decorations.pathreplacing, arrows.meta, fit, shapes.geometric}
\usepackage{amssymb}
\newcommand\dboxed[1]{\dbox{\ensuremath{#1}}}
\usepackage{amsmath} 
\usepackage{dashbox}
\usepackage{wasysym}
\usetikzlibrary{positioning}

\DeclareMathSymbol{\shortminus}{\mathbin}{AMSa}{"39}
\usetikzlibrary{positioning,chains}

\begin{document}

\title{Optimal Gradient Checkpointing for Sparse and Recurrent Architectures using Off-Chip Memory}

\author{\IEEEauthorblockN{Wadjih Bencheikh*}
\IEEEauthorblockA{\textit{Peter Gr{\"u}nberg Institut 15, } \\
\textit{FZ J{\"u}lich, ESI Algiers}\\
Aachen, Germany \\
bencheikh.wadjih@gmail.com}
\and
\IEEEauthorblockN{Jan Finkbeiner*}
\IEEEauthorblockA{\textit{Peter Gr{\"u}nberg Institut 15, } \\
\textit{FZ J{\"u}lich, RWTH Aachen}\\
Aachen, Germany \\
j.finkbeiner@fz-juelich.de}
\and
\IEEEauthorblockN{Emre Neftci}
\IEEEauthorblockA{\textit{Peter Gr{\"u}nberg Institut 15,} \\
\textit{FZ J{\"u}lich, RWTH Aachen}\\
Aachen, Germany \\
e.neftci@fz-juelich.de}
}

\maketitle

\begin{abstract}
Recurrent neural networks (RNNs) are valued for their computational efficiency and reduced memory requirements on tasks involving long sequence lengths but require high memory-processor bandwidth to train.
Checkpointing techniques can reduce the memory requirements by only storing a subset of intermediate states, the checkpoints, but are still rarely used due to the computational overhead of the additional recomputation phase. 
This work addresses these challenges by introducing memory-efficient gradient checkpointing strategies tailored for the general class of sparse RNNs and Spiking Neural Networks (SNNs). SNNs are energy efficient alternatives to RNNs thanks to their local, event-driven operation and potential neuromorphic implementation.  
We use the Intelligence Processing Unit (IPU) as an exemplary platform for architectures with distributed local memory. 
We exploit its suitability for sparse and irregular workloads to scale SNN training on long sequence lengths. 
We find that Double Checkpointing emerges as the most effective method, optimizing the use of local memory resources while minimizing recomputation overhead. 
This approach reduces dependency on slower large-scale memory access, enabling training on sequences over 10 times longer or 4 times larger networks than previously feasible, with only marginal time overhead. The presented techniques demonstrate significant potential to enhance scalability and efficiency in training sparse and recurrent networks across diverse hardware platforms, and highlights the benefits of sparse activations for scalable recurrent neural network training.
\end{abstract}


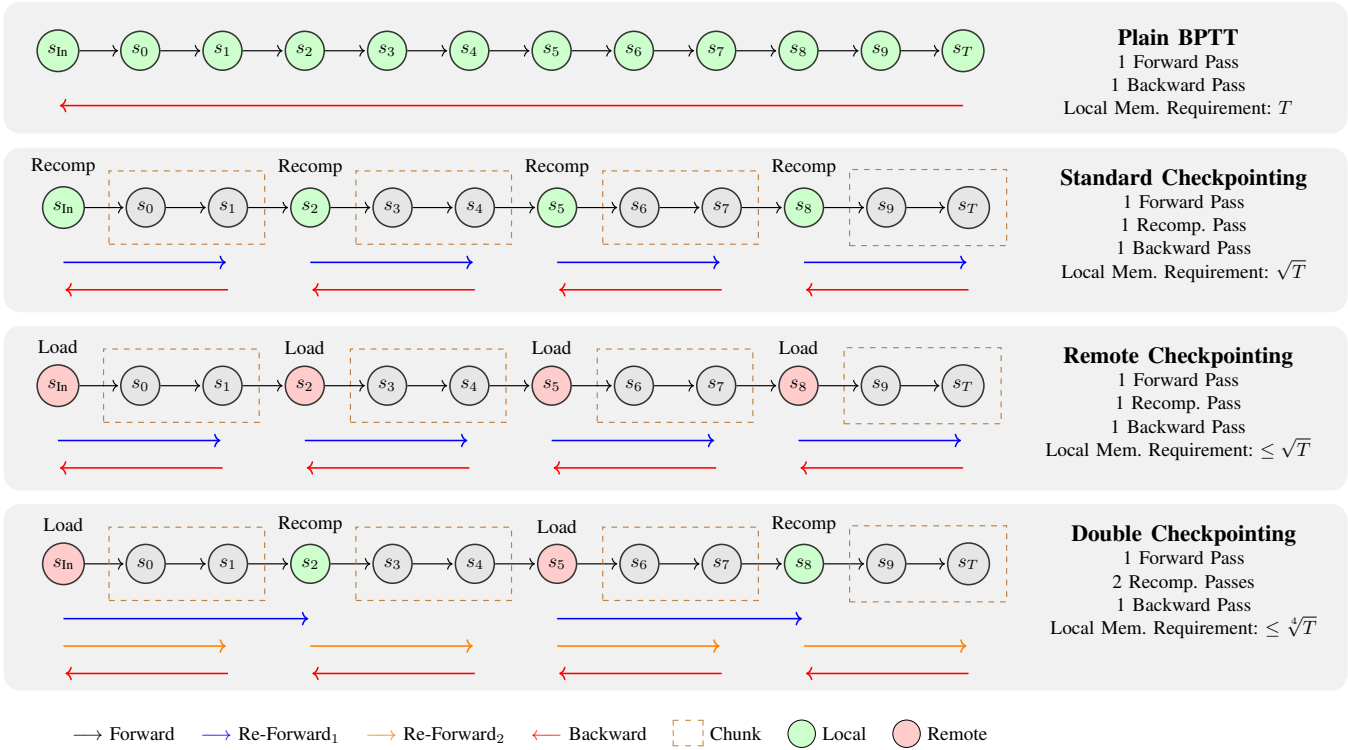
\begin{figure*}[t]
    \centering
    \resizebox{\textwidth}{!}{
        \input{figures/checkpointing_sketch}    
    }
    \caption{Comprehensive overview of the Backpropagation Through Time (BPTT) process, showcasing how intermediate states are handled and reconstructed. Execution traces for various gradient checkpointing strategies, including Standard, Chunk, Remote, and Double Checkpointing}
    \label{fig:enter-label}
\end{figure*}
\section{Introduction}
Sparse and recurrent neural networks represent a powerful paradigm in machine learning, with applications spanning time-series analysis, natural language processing, and event-driven computation \cite{schmidt_recurrent_2019}. Among these, Spiking Neural Networks (SNNs) have garnered significant interest for their biologically inspired, event-driven processing, which departs from the dense, synchronous data handling of traditional artificial neural networks (ANNs) \cite{lagani_spiking_2023}. SNNs communicate through sparse, asynchronous spike-based events, offering the potential for substantial improvements in computational efficiency and energy consumption \cite{ayasi_advancing_2024}. More broadly, sparse and recurrent architectures including SNNs and other networks with sparse activations are well-suited for scenarios involving irregular, event-driven, or sequence-based data processing \cite{narang_block-sparse_2017}.

However, the adoption of such networks has been constrained by hardware limitations. Most widely available hardware systems, such as Graphics Processing Units (GPUs), are optimized for dense matrix operations, making them less efficient for the sparse and recurrent workloads \cite{shi_efficient_2020}. This challenge is particularly pronounced in systems with limited high-speed local memory and reliance on slower, remote memory.

To address these limitations, we introduce memory-efficient training techniques designed for hardware architectures capable of handling sparse and recurrent computations. While SNNs serve as a focal application, our approach is generalizable to other sparse and recurrent networks, including those with sparse integer or floating-point activations. We leverage the Intelligence Processing Unit (IPU) as an ideal platform to showcase our methods, given its tile-based architecture and hierarchical memory system with its remote streaming memory. As such, the IPU is particularly well-suited to sparse and irregular workloads \cite{jia_dissecting_2019}. These features enable the IPU to efficiently manage the memory and computational demands of long-sequence and large model training, a task that typically requires significant resources \cite{shekofteh_performance_2023}.

Checkpointing allows to reduce the memory requirement during training by only storing a subset of intermediate states \cite{chen_training_2016}. For RNNs, this is typically implemented by storing the intermediate states at timesteps distanced at a certain intervals, the checkpoints, and dropping the states for timesteps inbetween \cite{gruslys2016memoryefficientbackpropagationtime}. During the backward phase, this requires the recomputation of the dropped states with an additional forward pass, which typically induces a roughly $30\%$ overhead in training time. 

Our contributions include the development of novel gradient checkpointing techniques tailored for sparse and recurrent architectures and for hardware architecures that do not feature high bandwidth memory (HBM), but some form of additional external, off-chip memory. Among these methods, Double Checkpointing emerges as the most effective, enabling sequence training lengths over 10 times longer and over 4 times bigger networks than previously feasible with minimal time overhead. This approach strategically balances memory efficiency and computational overhead by reducing reliance on slower memory systems and optimizing the use of local memory. Although the IPU serves as the primary platform for this work, our methods are broadly applicable across a wide range of hardware architectures. By exploiting sparse activations our techniques enable scalable and efficient training of sparse and recurrent networks, contributing to advancements in both hardware utilization and network scalability.
\section{Related Work}
The exploration of efficient neural network training and hardware optimization has been a prominent focus in AI research. Traditional approaches often leverage GPUs and TPUs \cite{10.1145/3140659.3080246}, but alternative architectures are being explored to improve the scalability and efficiency of models, particularly for sparse and recurrent networks. For example, Finkbeiner \textit{et al.} investigated the use of massively parallel MIMD architectures, such as IPUs, for sparse spiking neural networks (SNNs), achieving significant throughput gains over traditional GPUs \cite{finkbeiner_harnessing_2023}.

Another line of research has concentrated on improving learning mechanisms for SNNs. Bellec \textit{et al.} introduced LSNNs, which integrate neuronal adaptation for enhanced computing and learning capabilities, achieving computational performance comparable to LSTM networks \cite{bellec_long_2018}. Zenke and Vogels highlighted the robustness of surrogate gradient methods in enabling functional SNNs with sparse activity \cite{zenke_remarkable_2021}.

Memory efficiency is also critical, especially for recurrent and spiking networks. Chen \textit{et al.} proposed a sublinear memory algorithm for training deep networks, significantly reducing memory overhead with manageable computational trade-offs \cite{chen_training_2016}. Gruslys \textit{et al.} extended this idea to backpropagation through time (BPTT), reducing memory consumption while training RNNs on long sequences \cite{gruslys_memory-efficient_2016}. Singh \textit{et al.} further enhanced BPTT for SNNs by introducing time-skipping and activation-checkpointing techniques to address high memory requirements, achieving substantial speed and memory efficiency improvements \cite{singh_skipper_2022}.

The potential for hardware-specific optimization is also evident in FPGA-based systems, as demonstrated by Ramhorst \textit{et al.}, who developed a resource-aware structured pruning method tailored for FPGA, significantly reducing hardware resource usage \cite{ramhorst_fpga_2023}. Additionally, recent advances in adaptive memory strategies, such as dynamic programming-based memory optimization for neural networks, have allowed efficient utilization of computational resources without compromising training performance \cite{gruslys_memory-efficient_2016}.

Beyond hardware considerations, improving the training dynamics of SNNs is crucial. Advances like surrogate gradient learning and memory-optimized BPTT have laid the groundwork for enabling large-scale SNN models with practical training runtimes \cite{zenke_remarkable_2021, chen_training_2016}.

These studies provide the foundation for our research, where we explore advanced checkpointing techniques to improve both memory efficiency and computational
performance in SNN training on the IPU.
\begin{figure*}[t]
    \centering
    \includegraphics[width=\textwidth]{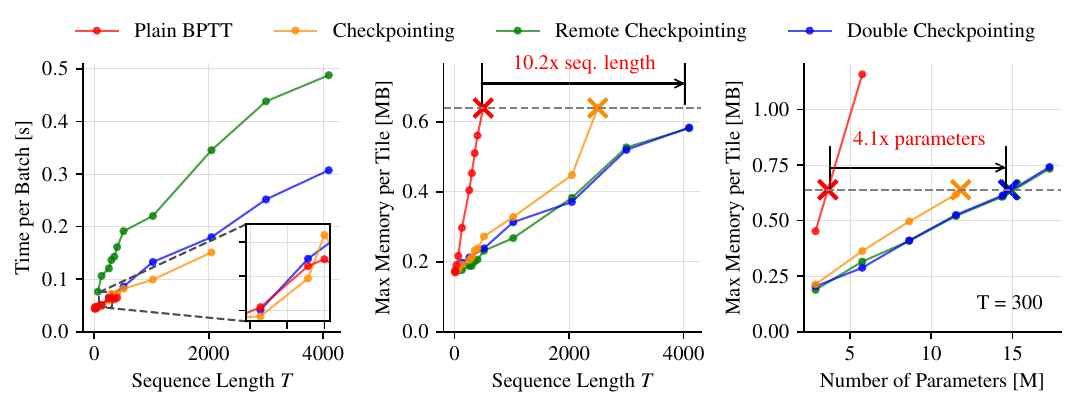}
    \label{fig:time_memory_results}
    \caption{Performance Comparison of Gradient Checkpointing Strategies Across Sequence Lengths and Model Sizes. Left: Time per batch as a function of sequence length, showing how Double Checkpointing maintains competitive training times compared to other methods, even for longer sequences. Middle: Peak local memory per tiles across sequence lengths, highlighting Double Checkpointing's ability to minimize memory usage while scaling. Right: Peak local memory per tiles as a function of model size, demonstrating Double Checkpointing's scalability and efficiency in handling larger models with T=300.}
\end{figure*}
\section{Methods}
This section outlines the core methodologies developed for training Spiking Neural Networks (SNNs) on the Graphcore Intelligence Processing Unit (IPU) \cite{jia2019dissectinggraphcoreipuarchitecture}, with a focus on memory efficiency and training time optimization. 

\subsection{BPTT based SNN training on the IPU}
The IPU, a massively parallel compute architecture with distributed local memory, is ideally suited for multiple data, multiple instruction (MIMD) workloads. Its programming paradigm is based on the Bulk Synchronous Parallel (BSP) model, which organizes computation into sequential supersteps. Each superstep consists of a local computation phase, a communication phase, and a barrier synchronization phase.

During local computation, each of the IPU’s cores has access to 624kB of dedicated SRAM, forming a ``tile'' with the core. This architecture enables highly efficient memory access due to its extremely low latency, comparable to L1 cache on GPUs. This feature, combined with the IPU's capability for efficient processing of unstructured sparsity, allows it to outperform traditional GPU and TPU architectures in scenarios involving recurrent neural networks with sparse connectivity and activations. Unlike GPUs, which require large, structured data packets (128 bytes minimum), the IPU can efficiently handle small data packets (8 bytes), making it especially well-suited for SNNs, which inherently feature both recurrence and sparsity.

Our base implementation builds upon the sparse representation outlined in \cite{finkbeiner_harnessing_2023}. This approach optimizes memory usage and accelerates calculations by using sparse spike representations and distributing neuron states and weights to dedicated tiles to minimize communication overhead. The models are trained using backpropagation through time (BPTT) using the surrogate gradient approach to account for the spikes discontinuity \cite{Zenke_2018, 8891809neftci}. This foundational implementation serves as the starting point for exploring advanced gradient checkpointing strategies, which we introduce in subsequent sections to further enhance memory efficiency and scalability of sparse RNN training, as demonstrated by SNN training on the IPU.
\begin{figure*}[t]
    \centering
    \includegraphics[width=\textwidth]{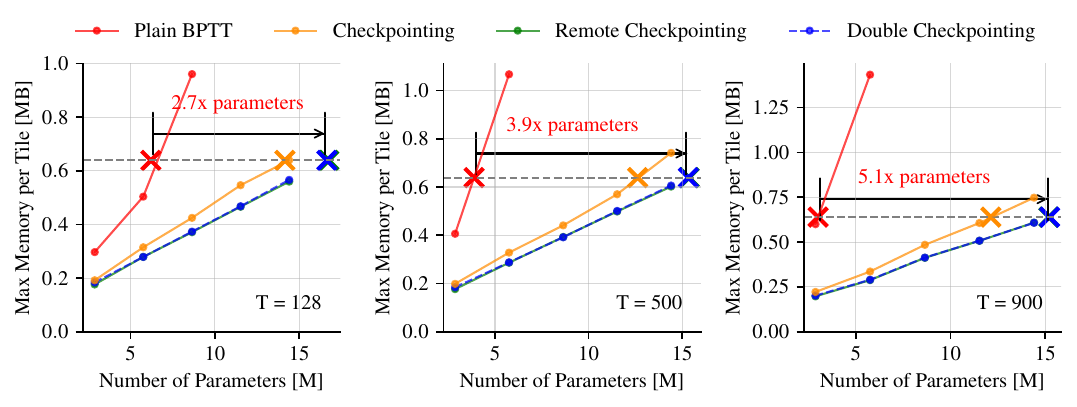}
    \caption{Memory Efficiency of Double Checkpointing Across Model Sizes and Configurations. The graphs compare the maximum peak local memory per tiles for Double Checkpointing versus the base implementation across varying sequence lengths (T) and batch sizes. Left: T=128 and batch\_size=120. Middle: T=500 and batch\_size=60. Right: T=900 and batch\_size=60}
    \label{fig:model_scaling}
\end{figure*}
\subsection{Standard Checkpointing}
Standard Checkpointing reduces memory usage by storing only a subset of neuron states during the forward pass. When performing the backward pass, these checkpoints are used to recompute intermediate states in smaller chunks. This avoids storing all intermediate states while maintaining computational feasibility.

On the IPU, the forward pass involves two main operations\cite{finkbeiner_harnessing_2023}: Matrix multiplication (MatMul) to calculate the next state, and Spike Generation function, which generate sparse spikes from states. While the Spike generation step is computationally expensive, saving the spikes avoids recomputing it during the backward pass. This optimization reduces the recomputation overhead, making the re-execution of the forward pass approximately $30\%$ of the original cost.

The chunk size parameter, \(\text{chunk\_size}\), determines the number of time-steps between checkpoints. The number of checkpoints is given by:
\begin{equation}
\text{nb\_checkpoints} = \frac{T}{\text{chunk\_size}}
\end{equation}
where $T$ is the sequence length.
The memory required for the Base Implementation is:
\begin{equation}
M_\text{Base} = M_{\text{s}} \cdot T + M_{\text{others}}
\end{equation}
where $M_\text{others}$ accounts for non-optimizable tensors such as spiking activations and synaptic weights, and $M_{\text{s}}$ represents the memory required for a single state:
\begin{equation}
M_{\text{s}} = \mathcal{O}(\text{batch\_size} \cdot \text{num\_neurons})
\end{equation}
%
With Standard Checkpointing across time, memory consumption becomes:
\begin{equation}
M_\text{Standard} = M_{\text{s}} \cdot (\text{chunk\_size} + \text{nb\_checkpoints}) + M_{\text{others}}
\end{equation}
Since we only save $\text{nb\_checkpoints}$ checkpoints, in addition to allocating a chunk of $\text{chunk\_size}$ state to save intermediate values between checkpoints. 
However, recomputation increases the time complexity by the time of the additional recomputation forward pass $T_\text{refwd}$:
\begin{equation}
T_\text{Standard} = T_\text{fwd} + T_\text{bwd} + T_\text{refwd}
\end{equation}
where $T_\text{fwd}$ and $T_\text{bwd}$ represent forward and backward pass times.
Since operations related to sparse representation generation or event communication are avoided during recomputation, $T_\text{refwd}$ constitutes less than $10\%$ of the total forward pass computation for the implementation on the IPU, highlighting the efficiency of Standard Checkpointing with sparse activations.

\subsection{Remote Checkpointing}
Due to the limited and expensive on-chip memory on most hardware architectures, most architectures introduce additional off-chip memory. The IPU features fast on-tile memory and slower, larger so called streaming memory. Remote Checkpointing leverages this architecture by offloading checkpoints to the external streaming memory, reducing on-tile memory consumption and enabling larger models and sequence lengths. However, access to off-chip memory introduces data transfer and synchronization overheads.

The memory requirement for Remote Checkpointing consequently loses its dependence on the number of checkpoints stored in local, on-chip memory:
\begin{equation}
M_\text{Remote} = M_{\text{s}} \cdot \left( \text{chunk\_size} + 1 \right) + M_{\text{others}}, 
\end{equation}
however, for the time complexity, the additional data transfer and synchronization times must be considered:
\begin{equation}
T_\text{Remote} = T_\text{Standard} + 2 \cdot (T_{\text{c}} + T_{\text{s}}) \cdot \text{nb\_checkpoints} ,
\end{equation}
where \( \text{T}_{\text{c}} \) is the time to transfer a single checkpoint to and from streaming memory, and $T_{\text{s}}$ is the average synchronization time. While Remote Checkpointing excels in memory efficiency, excessive data transfers can diminish its time efficiency.
\begin{figure*}[t]
    \centering
        \includegraphics[width=\textwidth]{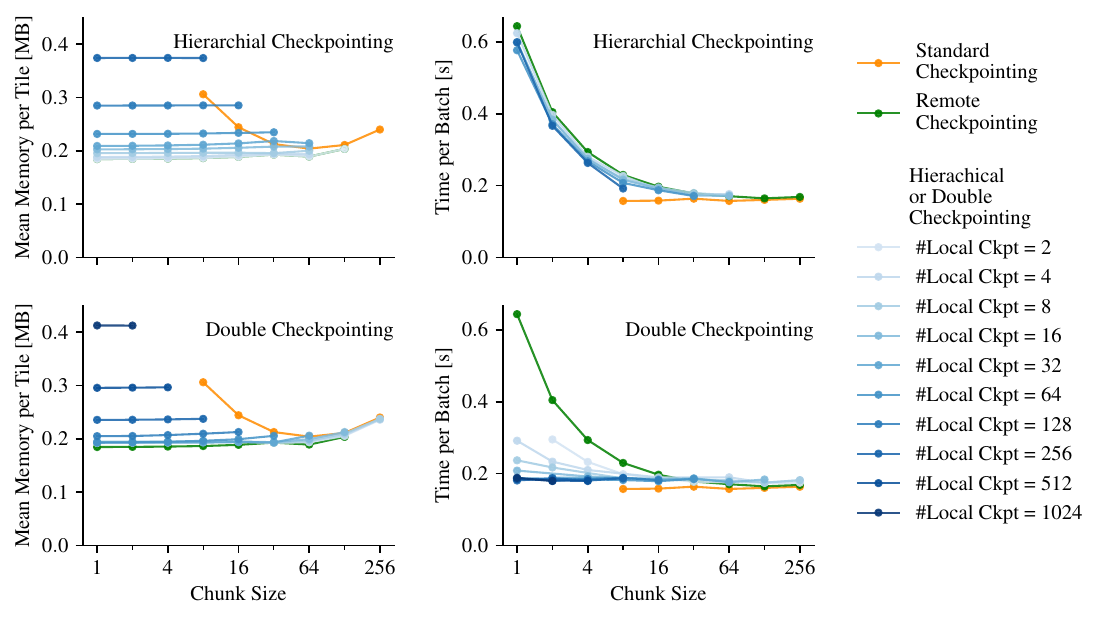}
    \caption{Hyperparameter Study of Double and Hierarchical Checkpointing. This figure illustrates the impact of varying chunk size and the number of local checkpoints on training time (right) and memory usage (left) for Double Checkpointing (bottom) and Hierarchical Checkpointing (top).}
    \label{fig:chunk_sweep}
\end{figure*}
\subsection{Hierarchical Checkpointing}
Hierarchical Checkpointing combines the benefits of Standard and Remote Checkpointing by fetching batches of checkpoints instead of individual ones. This minimizes data transfers and synchronization costs while optimizing memory usage.
Consequently, the memory requirement scales similar to the Standard Checkpointing case, however, with the number of local checkpoints $\text{nb\_local} \le \text{nb\_checkpoints}$:
\begin{equation}
M_{\text{Hier}} = M_{\text{s}} \cdot (\text{chunk\_size} + \text{nb\_local}) + M_{\text{others}} .
\end{equation}
The time complexity is:
\begin{equation}
T_{\text{Hier}} = T_{\text{Standard}} + (T_{\text{c}} \cdot \text{nb\_local} + T_{\text{s}}) \cdot N_\text{c} ,
\end{equation}
where \( N_\text{c} \), the number of communications with streaming memory, is defined as:
\begin{equation}
N_\text{c} = 2 \cdot \left(\frac{\text{nb\_checkpoints}}{\text{nb\_local}} - \text{nb\_local}\right) .
\end{equation}
Here, \( N_\text{c} \) is derived based on the hierarchical checkpointing strategy of fetching multiple checkpoints at once. Instead of retrieving remote checkpoints one by one, \( \text{nb\_local} \) checkpoints are read together in a single communication operation, which significantly reduces the number of synchronization steps required. The term \( \frac{\text{nb\_checkpoints}}{\text{nb\_local}} \) represents the number of batches of checkpoints fetched from remote memory. Since each batch requires communication for retrieval, this reduces the overall communication overhead. Furthermore, the last \( \text{nb\_local} \) checkpoints are directly stored locally, eliminating the need for further synchronization, which justifies the subtraction of \( \text{nb\_local} \) in the equation.  

Hierarchical Checkpointing strikes a balance between memory and computational overhead, making it well-suited for larger models with moderate sequence lengths.
\subsection{Double Checkpointing}
Double Checkpointing extends the concept of Hierarchical Checkpointing by introducing a two-tier checkpointing system. Remote checkpoints are complemented by intermediate local checkpoints between pairs of remote checkpoints, reducing frequent remote memory accesses. In contrast to Hierachical checkpointing, the local checkpoints are not loaded from remote memory as one block, but only one remote checkpoint is loaded from remote memory and the local checkpoints are recomputed with an additional recomputation forward pass. Therefore, the memory requirements are similar to the Hierachical Checkpointing case, however for the time complexity, time for remote data access and transfer is traded off for an additional recomputation pass:
\begin{align}
M_\text{Double} &= M_{\text{s}} \cdot (\text{chunk\_size} + \frac{\text{remote\_chunk\_size}}{\text{chunk\_size}}) + M_{\text{others}} , \\
T_\text{Double} &= T_\text{Standard} + T_\text{refwd} + ( T_{\text{c}} + T_{\text{s}}) \cdot N_\text{c} ,
\end{align}
where $N_\text{c}$, the number of communications with streaming memory (read and write), is:
\begin{equation}
N_\text{c} = 2 \cdot \frac{T}{\text{remote\_chunk\_size}} .
\end{equation}
By efficiently distributing memory and reducing communication overhead, Double Checkpointing enables scaling to both longer sequences and larger models, with minimal time penalties.
\begin{figure*}[t]
    \centering
        \includegraphics[width=\textwidth]{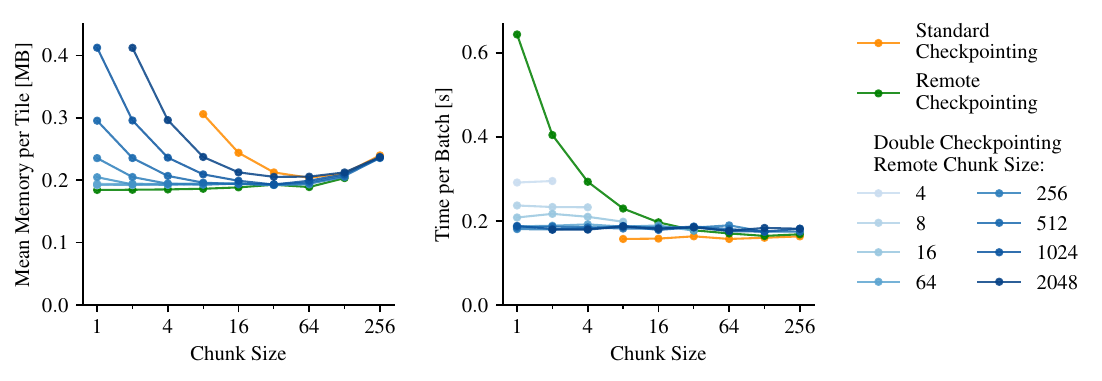}
    \caption{Optimization of Remote Chunk Size for Double Checkpointing. Exploring the relationship between remote chunk size and the performance of Double Checkpointing, balancing memory efficiency and computational time. Left: Mean local memory peak per tile as a function of remote chunk size. Smaller remote chunk sizes reduce local memory usage but require more frequent synchronization, leading to inefficiencies. Right: Time per batch as a function of remote chunk size. Larger remote chunk sizes minimize synchronization overhead, resulting in faster training times but higher memory usage.}
    \label{fig:chunk_sweep_double}
\end{figure*}
\section{Results}
We compared the performance of different checkpointing strategies on the  IPU by testing the maximum sequence length they could handle while maintaining acceptable training times. The experiments were conducted using the sparse spiking neural network on the Graphcore IPU with $3$ layers and $2$ neurons per tile. The batch size was set to $120$ for all experiments.

The results, shown in Figure \ref{fig:time_memory_results} (Left), reveal that the Base Implementation could only fit sequences of length $T=400$, making it unsuitable for large-scale tasks. On the other hand, Standard Checkpointing demonstrated the ability to scale up to sequences of length $2000$, with only a minimal increase in training time of approximately $5\%$.

Remote Checkpointing and Double Checkpointing strategies were able to scale up to sequences of length $T=4000$. However, the communication overhead for Remote Checkpointing caused noticeable time penalties, increasing training time by more then $50\%$. Double Checkpointing, however, achieved this scalability without significant time increases, making it the most balanced approach in terms of both memory usage and time efficiency.

In addition to sequence length scalability, we also evaluated the ability of these checkpointing strategies to handle larger model sizes. For this, we tried different sequence lengths (T) and batch sizes Figure \ref{fig:model_scaling} and in order to scale the network size, we changed the number of neurons per tile. Remarkably, both Double Checkpointing and Remote Checkpointing enabled the training of networks more than four times larger than what was achievable with the Base Implementation for $T=300$ in Figure \ref{fig:time_memory_results} (right), a typical sequence length for applications and datasets \cite{9311226-shd, orchard-2015-nmnist}. This significant increase in model size, combined with the ability to scale sequence lengths, underscores the effectiveness of these memory-efficient checkpointing strategies. Double Checkpointing, in particular, provided the best overall performance, offering both the scalability of longer sequences and the capacity to handle larger models without substantial time penalties.

These results highlight the scalability advantages of memory-efficient checkpointing strategies, particularly Double Checkpointing, which provides the best overall performance for long sequences and larger networks. 

\section{Hyperparameter Study}
In this study, we evaluated the performance of various checkpointing strategies by analyzing key hyperparameters. Since Standard Checkpointing is the most computationally efficient method and Remote Checkpointing is optimal for memory usage, we used these strategies as baselines for comparison with more complex methods. For Standard Checkpointing, theoretical analyses suggests that using \(\text{chunk\_size} = \sqrt{T}\), where \(T\) is the sequence length, strikes the best balance between memory and computation~\cite{gruslys_memory-efficient_2016}. Experimental results shown in Fig. \ref{fig:chunk_sweep} confirmed this, demonstrating that $\text{chunk\_size} = \sqrt{T} = 64$, for the given $T=4096$, optimally reduces memory usage while training times are unaffected by the choice of chunk size. In the case of Remote Checkpointing, memory consumption is minimized by selecting the smallest possible \(\text{chunk\_size}\). However, smaller chunk sizes increase communication overhead with streaming memory, creating a trade-off between memory requirements and time. Practical experiments revealed that the optimal $\text{chunk\_size}$ is the largest size that can fit in local memory, balancing memory efficiency with reduced communication costs.

We further investigated Hierarchical Checkpointing by varying the number of local checkpoints. Figure \ref{fig:chunk_sweep} (top) shows the results of that analysis. While increasing the number of local checkpoints effectively reduces memory usage, it does not significantly improve training time compared to Remote Checkpointing. 
While increasing the number of local checkpoints effectively increases local memory usage, it does not significantly improve training time compared to Remote Checkpointing. This inefficiency arises because the limited memory bandwidth, rather than synchronization, becomes the primary bottleneck on the IPU Streaming memory. Hierarchical Checkpointing mitigates slowdown caused by synchronization, but since synchronization overhead is not a dominant factor on the IPU, the benefits are limited.
For Double Checkpointing (Fig. \ref{fig:chunk_sweep}) (bottom) the memory requirement shows a similar trend as the Hierachical Checkpointing approach. However, while the training time for Hierachical Checkpointing did not improve for increasing number of local checkpoints, it does for Double Checkpointing. As a result, for Double Checkpointing configurations can be found that show low local memory requirements close to those of Remote Checkpointing while almost being as fast as Standard Checkpointing. The difference in results between the Hierachical and Double Checkpointing approach highlight that fast recomputation can be preferable over accessing and loading memory from external memory for checkpointing applications. 

Figure \ref{fig:chunk_sweep_double} shows the results where Double Checkpointing was additionally evaluated by varying the $\text{remote\_chunk\_size}$. Larger $\text{remote\_chunk\_size}$ reduced the number of remote checkpoints, which decreased communication time with streaming memory and improved training speeds. However, this came at the cost of increased memory usage due to the need for more local checkpoints. For a sequence length of $T = 4096$, the optimal configuration was found to be $\text{remote\_chunk\_size}$ between $64$ and $256$, which provided the best trade-off between memory efficiency and computational overhead, achieving a balance that minimized memory usage while maintaining reasonable training times. After fixing the $\text{remote\_chunk\_size}$ theoretical analysis suggests that using $\text{chunk\_size} = \sqrt{\text{remote\_chunk\_size}}$ is optimal which Fig. \ref{fig:chunk_sweep_double} (left) demonstrates experimentally.

\section{Discussion}
In this work, we demonstrated the potential of various checkpointing techniques for sparse and recurrent neural networks to alleviate the high memory requirements for back propagation trough time training, that naively scale linearly with the sequence length. Hereby, we specifically focus on hardware architectures with significant local, on-chip memory and without high bandwidth memory (HBM), but additional external, off-chip memory with limited bandwidth.  
Our checkpointing techniques exploit the sparsity of activations by storing the sparse activations of the full sequence in the local memory and only recomputing the internal states during checkpointing. Due to the drastically accelerated forward pass during recomputation we achieve almost the training time as without checkpointing. Additionally, we propose a Double Checkpoiting technique to even further reduce both the local memory requirements as well as the memory bandwidth requirements of the external memory. Using our novel checkpointing technique we demonstrate the training of spiking neural networks, a form of recurrent neural network with sparse and binary activations, for more than $10\,\times$ longer sequence lengths or $5\,\times$ larger networks at minimal training time overhead compared to plain backpropagation through time. 
Future work will require to expand the scope to explore the impact of these strategies on different types of neural network architectures and in various computational environments. Especially the the Double Checkpoint technique could enable long-sequence training even for purely on-chip memory scenarios due to it's drastically reduced memory requirement, $\mathcal{O}(\sqrt[4]{T})$, for choices of hyperparameters that minimize memory requirements. 

\section*{Acknowledgment}
This work was sponsored by the Federal Ministry of
Education, Germany BMBF under grants no. 16ME0398K,
16ME0399, 01IS22094E; and Neurosys as part of the initiative ”Cluster4Future” funded by the Federal Ministery of Education and Research BMBF (03ZU1106CB). The authors gratefully acknowledge computing time on the supercomputer JURECA (Jülich Supercomputing Centre 2021) at Forschungszentrum Jülich, providing access to Graphcore IPUs (as part of the JURECA DC Evaluation Platform).

\bibliographystyle{abbrv}
\bibliography{bibliography}

\end{document}

%% file: figures/checkpointing_sketch.tex
\begin{tikzpicture}[->,shorten >=1pt,auto,node distance=1.5cm,semithick]
    \tikzstyle{state}=[circle,draw=black!80,minimum size=0.6cm,fill=black!10,thick]  
    \tikzstyle{bbox}=[fill=gray!11, rounded corners=3mm, minimum width=24.5cm]

  \node[bbox, minimum height=2.4cm] (BBox) at (0, 0) {};

  \node[state,fill=green!20, anchor=center, xshift=1cm, yshift=-0.9cm] (Sin) at (BBox.north west)  {$s_\text{In}$};
  \node[state,fill=green!20] (S0) [right of=Sin] {$s_0$};
  \node[state,fill=green!20] (S1) [right of=S0] {$s_1$};
  \node[state,fill=green!20] (S2) [right of=S1] {$s_2$};
  \node[state,fill=green!20] (S3) [right of=S2] {$s_3$};
  \node[state,fill=green!20] (S4) [right of=S3] {$s_4$};
  \node[state,fill=green!20] (S5) [right of=S4] {$s_5$};
  \node[state,fill=green!20] (S6) [right of=S5] {$s_6$};
  \node[state,fill=green!20] (S7) [right of=S6] {$s_7$};
  \node[state,fill=green!20] (S8) [right of=S7] {$s_8$};
  \node[state,fill=green!20] (S9) [right of=S8] {$s_9$};
  \node[state,fill=green!20] (S10) [right of=S9] {$s_T$};

    \node[anchor=north, xshift=4.2cm, yshift=.25cm, align=center] (Desc) at (S10.north west) {\large\textbf{Plain BPTT}\\ 1 Forward Pass\\1 Backward Pass\\Local Mem. Requirement: $T$};
  
  \path (Sin) edge node {} (S0);
  \path (S0) edge node {} (S1);
  \path (S1) edge node {} (S2);
  \path (S2) edge node {} (S3);
  \path (S3) edge node {} (S4);
  \path (S4) edge node {} (S5);
  \path (S5) edge node {} (S6);
  \path (S6) edge node {} (S7);
  \path (S7) edge node {} (S8);
  \path (S8) edge node {} (S9);
  \path (S9) edge node {} (S10);

  \coordinate (start) at ([yshift=-1cm]Sin);
  \coordinate (end) at ([yshift=-1cm]S10);
  \draw[thick,->, red] (end) -- (start);

  \node[bbox, minimum height=3cm, below=1.45cm of S0] (BBox) at (0, 0) {};

  \node[state,fill=green!20, anchor=center, xshift=1.1cm, yshift=-1.1cm] (Sin) at (BBox.north west)  {$s_\text{In}$};
  \node[state] (S0) [right of=Sin] {$s_0$};
  \node[state] (S1) [right of=S0] {$s_1$};
  \node[state,fill=green!20] (S2) [right of=S1] {$s_2$};
  \node[state] (S3) [right of=S2] {$s_3$};
  \node[state] (S4) [right of=S3] {$s_4$};
  \node[state,fill=green!20] (S5) [right of=S4] {$s_5$};
  \node[state] (S6) [right of=S5] {$s_6$};
  \node[state] (S7) [right of=S6] {$s_7$};
  \node[state,fill=green!20] (S8) [right of=S7] {$s_8$};
  \node[state] (S9) [right of=S8] {$s_9$};
  \node[state] (S10) [right of=S9] {$s_T$};

    \node[anchor=north, xshift=4.2cm, yshift=.55cm, align=center] (Desc) at (S10.north west) {\large\textbf{Standard Checkpointing}\\ 1 Forward Pass\\1 Recomp. Pass\\1 Backward Pass\\Local Mem. Requirement: $\sqrt{T}$};

    \node[above=0.7mm of Sin] {Recomp};
    \node[above=0.7mm of S2] {Recomp};
    \node[above=0.7mm of S5] {Recomp};
    \node[above=0.7mm of S8] {Recomp};
  
  \node[fit={(S9) (S10)}, draw=brown, dashed, inner sep=0.3cm] (box) {};
  
  \node[fit={(S0) (S1)}, draw=brown, dashed, inner sep=0.3cm] (box) {};
  
  \node[fit={(S3) (S4)}, draw=brown, dashed, inner sep=0.3cm] (box) {};
  
  \node[fit={(S6) (S7)}, draw=brown, dashed, inner sep=0.3cm] (box) {};
  
  \path (Sin) edge node {} (S0);
  \path (S0) edge node {} (S1);
  \path (S1) edge node {} (S2);
  \path (S2) edge node {} (S3);
  \path (S3) edge node {} (S4);
  \path (S4) edge node {} (S5);
  \path (S5) edge node {} (S6);
  \path (S6) edge node {} (S7);
  \path (S7) edge node {} (S8);
  \path (S8) edge node {} (S9);
  \path (S9) edge node {} (S10);
  
  

  \coordinate (start) at ([yshift=-1cm]S8);
  \coordinate (end) at ([yshift=-1cm]S10);
  \draw[thick,->, blue] (start) -- (end);
  
  \coordinate (start) at ([yshift=-1.5cm]S8);
  \coordinate (end) at ([yshift=-1.5cm]S10);
  \draw[thick,->, red] (end) -- (start);

  \coordinate (start) at ([yshift=-1cm]S5);
  \coordinate (end) at ([yshift=-1cm]S7);
  \draw[thick,->, blue] (start) -- (end);
  
  \coordinate (start) at ([yshift=-1.5cm]S5);
  \coordinate (end) at ([yshift=-1.5cm]S7);
  \draw[thick,->, red] (end) -- (start);

  \coordinate (start) at ([yshift=-1cm]S2);
  \coordinate (end) at ([yshift=-1cm]S4);
  \draw[thick,->, blue] (start) -- (end);
  
  \coordinate (start) at ([yshift=-1.5cm]S2);
  \coordinate (end) at ([yshift=-1.5cm]S4);
  \draw[thick,->, red] (end) -- (start);
  
  \coordinate (start) at ([yshift=-1cm]Sin);
  \coordinate (end) at ([yshift=-1cm]S1);
  \draw[thick,->, blue] (start) -- (end);
  
  \coordinate (start) at ([yshift=-1.5cm]Sin);
  \coordinate (end) at ([yshift=-1.5cm]S1);
  \draw[thick,->, red] (end) -- (start);

  \node[bbox, minimum height=3cm, below=4.7cm of S0] (BBox) at (0, 0) {};

  \node[state,fill=red!20, anchor=center, xshift=1cm, yshift=-1.1cm] (Sin) at (BBox.north west)  {$s_\text{In}$};
  \node[state] (S0) [right of=Sin] {$s_0$};
  \node[state] (S1) [right of=S0] {$s_1$};
  \node[state,fill=red!20] (S2) [right of=S1] {$s_2$};
  \node[state] (S3) [right of=S2] {$s_3$};
  \node[state] (S4) [right of=S3] {$s_4$};
  \node[state,fill=red!20] (S5) [right of=S4] {$s_5$};
  \node[state] (S6) [right of=S5] {$s_6$};
  \node[state] (S7) [right of=S6] {$s_7$};
  \node[state,fill=red!20] (S8) [right of=S7] {$s_8$};
  \node[state] (S9) [right of=S8] {$s_9$};
  \node[state] (S10) [right of=S9] {$s_T$};

    \node[anchor=north, xshift=4.2cm, yshift=.55cm, align=center] (Desc) at (S10.north west) {\large\textbf{Remote Checkpointing}\\ 1 Forward Pass\\1 Recomp. Pass\\ 1 Backward Pass\\Local Mem. Requirement: $\leq \sqrt{T}$};

    \node[above=0.7mm of Sin] (descs2) {Load};
    \node[above=0.7mm of S2] (descs2) {Load};
    \node[above=0.7mm of S5] (descs2) {Load};
    \node[above=0.7mm of S8] (descs2) {Load};

  \node[fit={(S9) (S10)}, draw=brown, dashed, inner sep=0.3cm] (box) {};
  
  \node[fit={(S0) (S1)}, draw=brown, dashed, inner sep=0.3cm] (box) {};
  
  \node[fit={(S3) (S4)}, draw=brown, dashed, inner sep=0.3cm] (box) {};
  
  \node[fit={(S6) (S7)}, draw=brown, dashed, inner sep=0.3cm] (box) {};
  
  \path (Sin) edge node {} (S0);
  \path (S0) edge node {} (S1);
  \path (S1) edge node {} (S2);
  \path (S2) edge node {} (S3);
  \path (S3) edge node {} (S4);
  \path (S4) edge node {} (S5);
  \path (S5) edge node {} (S6);
  \path (S6) edge node {} (S7);
  \path (S7) edge node {} (S8);
  \path (S8) edge node {} (S9);
  \path (S9) edge node {} (S10);
  
  \coordinate (start) at ([yshift=-1cm]S8);
  \coordinate (end) at ([yshift=-1cm]S10);
  \draw[thick,->, blue] (start) -- (end);
  
  \coordinate (start) at ([yshift=-1.5cm]S8);
  \coordinate (end) at ([yshift=-1.5cm]S10);
  \draw[thick,->, red] (end) -- (start);

  \coordinate (start) at ([yshift=-1cm]S5);
  \coordinate (end) at ([yshift=-1cm]S7);
  \draw[thick,->, blue] (start) -- (end);
  
  \coordinate (start) at ([yshift=-1.5cm]S5);
  \coordinate (end) at ([yshift=-1.5cm]S7);
  \draw[thick,->, red] (end) -- (start);

  \coordinate (start) at ([yshift=-1cm]S2);
  \coordinate (end) at ([yshift=-1cm]S4);
  \draw[thick,->, blue] (start) -- (end);
  
  \coordinate (start) at ([yshift=-1.5cm]S2);
  \coordinate (end) at ([yshift=-1.5cm]S4);
  \draw[thick,->, red] (end) -- (start);
  
  \coordinate (start) at ([yshift=-1cm]Sin);
  \coordinate (end) at ([yshift=-1cm]S1);
  \draw[thick,->, blue] (start) -- (end);
  
  \coordinate (start) at ([yshift=-1.5cm]Sin);
  \coordinate (end) at ([yshift=-1.5cm]S1);
  \draw[thick,->, red] (end) -- (start);

  \node[bbox, minimum height=3.4cm, below=7.95cm of Sin] (BBox) at (0, 0) {};

  \node[state,fill=red!20, anchor=center, xshift=1.1cm, yshift=-1.1cm] (Sin) at (BBox.north west)  {$s_\text{In}$};
  \node[state] (S0) [right of=Sin] {$s_0$};
  \node[state] (S1) [right of=S0] {$s_1$};
  \node[state,fill=green!20] (S2) [right of=S1] {$s_2$};
  \node[state] (S3) [right of=S2] {$s_3$};
  \node[state] (S4) [right of=S3] {$s_4$};
  \node[state,fill=red!20] (S5) [right of=S4] {$s_5$};
  \node[state] (S6) [right of=S5] {$s_6$};
  \node[state] (S7) [right of=S6] {$s_7$};
  \node[state,fill=green!20] (S8) [right of=S7] {$s_8$};
  \node[state] (S9) [right of=S8] {$s_9$};
  \node[state] (S10) [right of=S9] {$s_T$};


    \node[anchor=north, xshift=4.2cm, yshift=.55cm, align=center] (Desc) at (S10.north west) {\large\textbf{Double Checkpointing}\\ 1 Forward Pass\\2 Recomp. Passes\\ 1 Backward Pass\\Local Mem. Requirement: $\leq \sqrt[4]{T}$}; 

    \node[above=0.7mm of Sin] {Load};
    \node[above=0.7mm of S2] {Recomp};
    \node[above=0.7mm of S5] {Load};
    \node[above=0.7mm of S8] {Recomp};
  
  \path (Sin) edge node {} (S0);
  \path (S0) edge node {} (S1);
  \path (S1) edge node {} (S2);
  \path (S2) edge node {} (S3);
  \path (S3) edge node {} (S4);
  \path (S4) edge node {} (S5);
  \path (S5) edge node {} (S6);
  \path (S6) edge node {} (S7);
  \path (S7) edge node {} (S8);
  \path (S8) edge node {} (S9);
  \path (S9) edge node {} (S10);
  
  \coordinate (start) at ([yshift=-1cm]S5);
  \coordinate (end) at ([yshift=-1cm]S8);
  \draw[thick,->, blue] (start) -- (end);

  \coordinate (start) at ([yshift=-1.5cm]S8);
  \coordinate (end) at ([yshift=-1.5cm]S10);
  \draw[thick,->, orange] (start) -- (end);
  
  \node[fit={(S9) (S10)}, draw=brown, dashed, inner sep=0.3cm] (box) {};
  
  \coordinate (start) at ([yshift=-2cm]S10);
  \coordinate (end) at ([yshift=-2cm]S8);
  \draw[thick,->, red] (start) -- (end);

  \coordinate (start) at ([yshift=-1cm]Sin);
  \coordinate (end) at ([yshift=-1cm]S2);
  \draw[thick,->, blue] (start) -- (end);
  
  \coordinate (start) at ([yshift=-1.5cm]S5);
  \coordinate (end) at ([yshift=-1.5cm]S7);
  \draw[thick,->, orange] (start) -- (end);
  
  \node[fit={(S6) (S7)}, draw=brown, dashed, inner sep=0.3cm] (box) {};
  
  \coordinate (start) at ([yshift=-2cm]S7);
  \coordinate (end) at ([yshift=-2cm]S5);
  \draw[thick,->, red] (start) -- (end);

  \coordinate (start) at ([yshift=-1.5cm]S2);
  \coordinate (end) at ([yshift=-1.5cm]S4);
  \draw[thick,->, orange] (start) -- (end);
  
  \node[fit={(S3) (S4)}, draw=brown, dashed, inner sep=0.3cm] (box) {};
  
  \coordinate (start) at ([yshift=-2cm]S4);
  \coordinate (end) at ([yshift=-2cm]S2);
  \draw[thick,->, red] (start) -- (end);

  \coordinate (start) at ([yshift=-1.5cm]Sin);
  \coordinate (end) at ([yshift=-1.5cm]S1);
  \draw[thick,->, orange] (start) -- (end);
  
  \node[fit={(S0) (S1)}, draw=brown, dashed, inner sep=0.3cm] (box) {};
  
  \coordinate (start) at ([yshift=-2cm]S1);
  \coordinate (end) at ([yshift=-2cm]Sin);
  \draw[thick,->, red] (start) -- (end);

  \node[fill=white, inner sep=0.5cm] (legend) at ($(S0) + (7, -3.1)$) {
      \textcolor{black}{\textbf{$\longrightarrow$}}  Forward \quad
      \textcolor{blue}{\textbf{$\longrightarrow$}}  $\text{Re-Forward}_1$  \quad
      \textcolor{orange}{\textbf{$\longrightarrow$}} $\text{Re-Forward}_2$ \quad
      \textcolor{red}{\textbf{$\longleftarrow$}}  Backward \quad
      \textcolor{brown}{$\dboxed{\textcolor{white}{X}}$}  Chunk \quad
      \raisebox{-0.15cm}{\tikz\node[fill=green!20, draw=black, circle, minimum size=0.5cm, inner sep=0pt, yshift=5cm] (node) {};} Local \quad
      \raisebox{-0.15cm}{\tikz\node[fill=red!20, draw=black, circle, minimum size=0.5cm, inner sep=0pt, yshift=5cm] (node) {};} Remote
  };
  
\end{tikzpicture}

%% file: main.bbl
\begin{thebibliography}{10}

\bibitem{ayasi_advancing_2024}
B.~Ayasi, Ã.~M. García-Vico, C.~J. Carmona, and M.~Saleh.
\newblock Advancing computational frontiers: Spiking neural networks
  in high-energy efficiency computing across diverse domains.
\newblock In A.~Alonso-Betanzos, B.~Guijarro-Berdiñas, V.~Bolón-Canedo,
  E.~Hernández-Pereira, O.~Fontenla-Romero, D.~Camacho, J.~R. Rabuñal,
  M.~Ojeda-Aciego, J.~Medina, J.~C. Riquelme, and A.~Troncoso, editors, {\em
  Advances in Artificial Intelligence}, pages 9--18. Springer Nature
  Switzerland, 2024.

\bibitem{bellec_long_2018}
G.~Bellec, D.~Salaj, A.~Subramoney, R.~Legenstein, and W.~Maass.
\newblock Long short-term memory and learning-to-learn in networks of spiking
  neurons.

\bibitem{chen_training_2016}
T.~Chen, B.~Xu, C.~Zhang, and C.~Guestrin.
\newblock Training deep nets with sublinear memory cost.

\bibitem{9311226-shd}
B.~Cramer, Y.~Stradmann, J.~Schemmel, and F.~Zenke.
\newblock The heidelberg spiking data sets for the systematic evaluation of
  spiking neural networks.
\newblock {\em IEEE Transactions on Neural Networks and Learning Systems},
  33(7):2744--2757, 2022.

\bibitem{finkbeiner_harnessing_2023}
J.~Finkbeiner, T.~Gmeinder, M.~Pupilli, A.~Titterton, and E.~Neftci.
\newblock Harnessing manycore processors with distributed memory for
  accelerated training of sparse and recurrent models.

\bibitem{gruslys_memory-efficient_2016}
A.~Gruslys, R.~Munos, I.~Danihelka, M.~Lanctot, and A.~Graves.
\newblock Memory-efficient backpropagation through time.

\bibitem{gruslys2016memoryefficientbackpropagationtime}
A.~Gruslys, R.~Munos, I.~Danihelka, M.~Lanctot, and A.~Graves.
\newblock Memory-efficient backpropagation through time, 2016.

\bibitem{jia_dissecting_2019}
Z.~Jia, B.~Tillman, M.~Maggioni, and D.~P. Scarpazza.
\newblock Dissecting the graphcore {IPU} architecture via microbenchmarking.

\bibitem{jia2019dissectinggraphcoreipuarchitecture}
Z.~Jia, B.~Tillman, M.~Maggioni, and D.~P. Scarpazza.
\newblock Dissecting the graphcore ipu architecture via microbenchmarking,
  2019.

\bibitem{10.1145/3140659.3080246}
N.~P. Jouppi, C.~Young, N.~Patil, D.~Patterson, G.~Agrawal, R.~Bajwa, S.~Bates,
  S.~Bhatia, N.~Boden, A.~Borchers, R.~Boyle, P.-l. Cantin, C.~Chao, C.~Clark,
  J.~Coriell, M.~Daley, M.~Dau, J.~Dean, B.~Gelb, T.~V. Ghaemmaghami,
  R.~Gottipati, W.~Gulland, R.~Hagmann, C.~R. Ho, D.~Hogberg, J.~Hu, R.~Hundt,
  D.~Hurt, J.~Ibarz, A.~Jaffey, A.~Jaworski, A.~Kaplan, H.~Khaitan,
  D.~Killebrew, A.~Koch, N.~Kumar, S.~Lacy, J.~Laudon, J.~Law, D.~Le, C.~Leary,
  Z.~Liu, K.~Lucke, A.~Lundin, G.~MacKean, A.~Maggiore, M.~Mahony, K.~Miller,
  R.~Nagarajan, R.~Narayanaswami, R.~Ni, K.~Nix, T.~Norrie, M.~Omernick,
  N.~Penukonda, A.~Phelps, J.~Ross, M.~Ross, A.~Salek, E.~Samadiani, C.~Severn,
  G.~Sizikov, M.~Snelham, J.~Souter, D.~Steinberg, A.~Swing, M.~Tan,
  G.~Thorson, B.~Tian, H.~Toma, E.~Tuttle, V.~Vasudevan, R.~Walter, W.~Wang,
  E.~Wilcox, and D.~H. Yoon.
\newblock In-datacenter performance analysis of a tensor processing unit.
\newblock {\em SIGARCH Comput. Archit. News}, 45(2):1–12, jun 2017.

\bibitem{lagani_spiking_2023}
G.~Lagani, F.~Falchi, C.~Gennaro, and G.~Amato.
\newblock Spiking neural networks and bio-inspired supervised deep learning: A
  survey.

\bibitem{narang_block-sparse_2017}
S.~Narang, E.~Undersander, and G.~Diamos.
\newblock Block-sparse recurrent neural networks.

\bibitem{8891809neftci}
E.~O. Neftci, H.~Mostafa, and F.~Zenke.
\newblock Surrogate gradient learning in spiking neural networks: Bringing the
  power of gradient-based optimization to spiking neural networks.
\newblock {\em IEEE Signal Processing Magazine}, 36(6):51--63, 2019.

\bibitem{orchard-2015-nmnist}
G.~Orchard, A.~Jayawant, G.~Cohen, and N.~V. Thakor.
\newblock {Converting static image datasets to spiking neuromorphic datasets
  using saccades}.
\newblock {\em Frontiers in Neuroscience}, 9, 11 2015.

\bibitem{ramhorst_fpga_2023}
B.~Ramhorst, V.~Loncar, and G.~A. Constantinides.
\newblock {FPGA} resource-aware structured pruning for real-time neural
  networks.
\newblock version: 2.

\bibitem{schmidt_recurrent_2019}
R.~M. Schmidt.
\newblock Recurrent neural networks ({RNNs}): A gentle introduction and
  overview.

\bibitem{shekofteh_performance_2023}
S.-K. Shekofteh, C.~Alles, N.~Kochendörfer, and H.~Fröning.
\newblock On performance analysis of graphcore {IPUs}: Analyzing squared and
  skewed matrix multiplication.

\bibitem{shi_efficient_2020}
S.~Shi, Q.~Wang, and X.~Chu.
\newblock Efficient sparse-dense matrix-matrix multiplication on {GPUs} using
  the customized sparse storage format.
\newblock In {\em 2020 {IEEE} 26th International Conference on Parallel and
  Distributed Systems ({ICPADS})}, pages 19--26, 2020.
\newblock {ISSN}: 2690-5965.

\bibitem{singh_skipper_2022}
S.~Singh, A.~Sarma, S.~Lu, A.~Sengupta, M.~T. Kandemir, E.~Neftci,
  V.~Narayanan, and C.~R. Das.
\newblock Skipper: Enabling efficient {SNN} training through
  activation-checkpointing and time-skipping.
\newblock In {\em 2022 55th {IEEE}/{ACM} International Symposium on
  Microarchitecture ({MICRO})}, pages 565--581, 2022.

\bibitem{Zenke_2018}
F.~Zenke and S.~Ganguli.
\newblock Superspike: Supervised learning in multilayer spiking neural
  networks.
\newblock {\em Neural Computation}, 30(6):1514–1541, June 2018.

\bibitem{zenke_remarkable_2021}
F.~Zenke and T.~P. Vogels.
\newblock The remarkable robustness of surrogate gradient learning for
  instilling complex function in spiking neural networks.
\newblock {\em Neural Computation}, 33(4):899--925, 2021.

\end{thebibliography}
